\documentclass[letterpaper, 10 pt, conference]{ieeeconf}  

\IEEEoverridecommandlockouts                              

\usepackage[font=small]{caption}
\usepackage{graphicx} 
\usepackage{tabularx}
\usepackage{times} 
\usepackage{amsmath} 
\usepackage{amssymb}  
\usepackage{comment}
\usepackage{url}
\usepackage{subfigure}
\usepackage{multirow}
\usepackage{colortbl}
\usepackage{booktabs}
\usepackage[table,xcdraw]{xcolor}
\usepackage[normalem]{ulem}
\usepackage{xcolor}
\usepackage{xspace}
\usepackage{pifont}

\makeatletter
\let\NAT@parse\undefined
\makeatother
\usepackage[pagebackref=false,breaklinks=true,letterpaper=true,colorlinks,bookmarks=false]{hyperref}

\definecolor{sparkvln_green}  {RGB}{ 46, 139,  87}
\definecolor{sparkvln_yellow} {RGB}{218, 165,  32}
\definecolor{sparkvln_gray}   {RGB}{120, 120, 128}
\definecolor{sparkvln_blue}   {RGB}{ 70, 110, 180}
\definecolor{sparkvln_red}     {RGB}{255, 127, 80}
\definecolor{sparkvln_orange}  {RGB}{255, 140, 0}

\usepackage{multirow}
\usepackage{xcolor}
\newcommand{\gc}[1]{\textcolor{green!50!black}{#1}}

\usepackage{xcolor}

\definecolor{sparkA}{HTML}{2FA8E0}
\definecolor{sparkB}{HTML}{2447B8}

\colorlet{spk1}{sparkA!100!sparkB}  
\colorlet{spk2}{sparkA!88!sparkB}   
\colorlet{spk3}{sparkA!75!sparkB}   
\colorlet{spk4}{sparkA!62!sparkB}   
\colorlet{spk5}{sparkA!50!sparkB}   
\colorlet{spk6}{sparkA!38!sparkB}   
\colorlet{spk7}{sparkA!25!sparkB}   
\colorlet{spk8}{sparkA!12!sparkB}   
\colorlet{spk9}{sparkA!0!sparkB}    

\newcommand{\sparkvln}{\textbf{%
\textcolor{spk1}{S}\textcolor{spk2}{P}\textcolor{spk3}{A}%
\textcolor{spk4}{R}\textcolor{spk5}{K}\textcolor{spk6}{-}%
\textcolor{spk7}{V}\textcolor{spk8}{L}\textcolor{spk9}{N}}}

\newcommand{\sparkvlnplain}{SPARK-VLN}

\newcommand{\spkw}[3]{\textcolor{#1}{\textbf{#2}}#3}

\title{\LARGE \bf
 Token-Wise Latent Streaming from Slow Reasoners to Fast \\ Planners for Dynamic Vision Language Navigation
}

\author{Tianshuai Hu$^{1}$, Yangyi Zhong$^{2}$, Zeying Gong$^{2}$, Lingdong Kong$^{3}$, Xiaodong Mei$^{1}$,\\ Guoyang Zhao$^{2}$, Xiaolu Liu$^{4}$, Song Wang$^{4}$, Rong Li$^{2}$,  and Junwei Liang$^{1,2*}$%
\thanks{$^{1}$ The Hong Kong University of Science and Technology. {\tt\footnotesize \{thuaj, xmeiab\}@connect.ust.hk}}%
\thanks{$^{2}$ The Hong Kong University of Science and Technology (Guangzhou). 
{\tt\footnotesize \{yzhong123, zgong313, rli335\}@connect.hkust-gz.edu.cn, junweiliang@hkust-gz.edu.cn}}%
\thanks{$^{3}$ National University of Singapore. {\tt\footnotesize lingdong.kong@u.nus.edu}}%
\thanks{$^{4}$ Zhejiang University {\tt\footnotesize \{xiaoluliu, songw\}@zju.edu.cn}}
\thanks{* Corresponding author.}
}

\begin{document}

\maketitle
\thispagestyle{empty}
\pagestyle{empty}

Vision-Language Navigation in dynamic, human-centric environments exposes a fundamental tension: linguistic reasoning is slow and  deliberative, whereas safe, socially compliant planning should be  instant and reactive. The resulting observation staleness is safety-critical: a maneuver chosen during inference can already  be unsafe by the time it executes. We observe that, long before a  VLM finishes its inference, its intermediate hidden states already  encode action-relevant intent. We propose \sparkvln{}, a dual-system framework for dynamic social VLN that streams the slow VLM reasoner's knowledge to a fast flow-matching expert planner throughout token generation, providing fresh and evolving guidance during inference. This design is realized by three modules: a Token-Wise Hidden Streamer that extracts intermediate hidden states along the token generation process, a Sequence-to-Slot Latent Bridge that projects them into fixed-size latent slots, and an Evolving Latent Conditioner that infuses them into the expert planner. We also introduce a human-centric  benchmark suite for dynamic social vision-language navigation that  keeps pedestrians and the robot active throughout inference and reports navigation success, social compliance, human collisions, and explicit staleness statistics. Across these settings, \sparkvlnplain{} improves navigation success and social compliance while sustaining inference efficiency. Webpage:~\url{https://hutslib.github.io/SPARK-VLN/}.
\begin{figure}[t]
\centering
\includegraphics[width=0.8\linewidth]{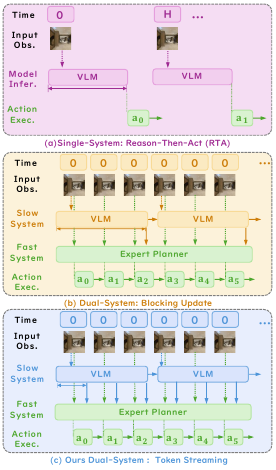}
\caption{\textbf{Comparison of VLN execution paradigms.}
\textbf{(a)~Single-System (Reason-Then-Act):} the VLM must finish full inference before emitting an action; the agent idles while the environment advances.
\textbf{(b)~Dual-System (Blocking Update):} the expert planner acts continuously, but receives guidance (yellow arrows) only after the VLM finishes, yielding a stale conditioning signal.
\textbf{(c)~Ours (Token Streaming):} \sparkvln{} streams hidden states to the planner throughout inference (blue arrows), providing progressively refined guidance.}
\label{fig:paradigm_comparison}
\vspace{-1em}
\end{figure}
\section{Introduction}
\label{sec:introduction}

Robots deployed in human-centric spaces are increasingly required to follow natural-language instructions while moving safely among dynamic pedestrians, reasoning over task semantics, scene context, and social norms. Vision-Language-Navigation~(VLN) models are a promising route to instruction following~\cite{zhang2024navid, cheng2025navila, zhang2024uni, wei2025streamvln, wei2026ground}. Deploying them in dynamic social spaces, however, exposes a fundamental tension: semantic reasoning is deliberative and slow, whereas safe social navigation demands instant reactions~\cite{hu2026navthinker, gong2026flux, gong2025cognition}.

The reasoning of multimodal large language models (MLLMs)~\cite{Qwen2.5-VL,li2024llama, lin2024vila,achiam2023gpt,chen2024internvl} is inherently slow, and while the model deliberates, the environment keeps evolving. The result is a perception-action gap we term \emph{observation staleness}. While static benchmarks hide it, dynamic environments make it severe: an action is computed from an observation that the world has already outrun, so a maneuver that was socially appropriate when inference began can become a proxemic violation, or even a collision, by the time it executes. Existing methods do not fully resolve this problem. Reason-then-act systems~\cite{zhang2024navid, cheng2025navila, zhang2024uni} ignore latency entirely and pause between actions. Real-Time Chunking (RTC) and continuous-inference pipelines~\cite{black2026real, black2025training, xie2026dynamicvla} overlap inference with execution to preserve motion smoothness, yet each chunk is computed from an observation the environment has already outrun~\cite{tang2025vlash, lu2026faster}. Dual-system designs~\cite{wei2026ground, wei2025streamvln, huang2026tic} decouple a slow reasoner from a fast planner to recover reactivity, but guidance is handed over only after the slow Vision-Language Model (VLM) finishes reasoning, and is already outdated by the time it steers the fast planner.

Compounding the issue, this failure mode is largely invisible under current evaluation. Mainstream VLN benchmarks~\cite{anderson2018vision, ku2020room, krantz2020beyond} are static and synchronous: the simulator freezes while the model thinks and advances only after an action is emitted, so inference latency never alters the outcome. As a result, observation staleness is neither penalized nor even measured, and methods that would fail in a live dynamic scene appear indistinguishable from methods that would not.

In this work, we revisit the dual-system paradigm, which preserves large-model reasoning while restoring reactivity. Our key observation is that autoregressive inference is incremental: long before a VLM finishes its answer, the hidden states of its intermediate tokens already encode cues such as directional intent and pedestrian context. Existing dual-system designs~\cite{wei2026ground, wei2025streamvln, huang2026tic} ignore this. They treat reasoning as a blocking call: they \textit{extract} guidance only from a finished chain, \textit{project} it into a single final latent for the downstream fast planner, and \textit{infuse} that into the planner once per slow cycle. As a result, throughout the long inference window the planner receives no fresh guidance at all, and by the time the guidance lands, the scene it was reasoned from has already changed. The bottleneck, then, lies not in the reasoning process but in its delivery: guidance is withheld until reasoning completes, even though usable knowledge emerges token by token along the way.

We therefore propose \sparkvln{}, a dual-system framework for dynamic social VLN in which a slow \emph{VLM reasoner} \spkw{spk1}{S}{treams} \spkw{spk2}{P}{rogressively} \spkw{spk3}{A}{ggregated} latent \spkw{spk4}{R}{easoning} \spkw{spk5}{K}{nowledge} to a fast \emph{flow-matching expert planner} throughout inference. Concretely, \sparkvlnplain{} rebuilds the extract--project--infuse pipeline into a token-wise streaming process. (i) Instead of extracting guidance only from a finished chain, the \textbf{Token-Wise Hidden Streamer} \textit{extracts} hidden states at intermediate tokens, so guidance becomes available while the VLM is still reasoning. (ii) The extracted hidden states, however, form an ever-growing sequence in the VLM's high-dimensional space, whereas the planner demands a fixed-size, low-dimensional condition: the \textbf{Sequence-to-Slot Latent Bridge} therefore \textit{projects} these hidden states into a compact set of latent slots via progressive aggregation with learnable queries, decoupling the planner's input from the VLM output length. (iii) To plan with both the evolving guidance and the current observation, the \textbf{Evolving Latent Conditioner} \textit{infuses} the up-to-date slots into the expert planner, which generates trajectories from the real-time observation together with the streamed guidance. As a result, the planner is steered by fresh semantic intent throughout the VLM's reasoning rather than only at its end.

To close the evaluation gap identified above, we further introduce a human-centric benchmark suite for dynamic social VLN that keeps both pedestrians and the robot active throughout model inference. The suite features: (1) two evaluation stages, an Idealized Dynamic Environment that pauses the simulator during inference and a Realistic Dynamic Environment where pedestrians keep moving as the model computes; (2) realistic trajectories, where goal-driven pedestrians avoid one another via Optimal Reciprocal Collision Avoidance (ORCA)~\cite{vanDenBerg2011ORCA}; (3) reasonable density and diverse social scenarios, with pedestrian counts scaled to scene area and safety-critical encounter patterns such as intersection, following, and corner; and (4) comprehensive metrics covering task completion, social compliance, and inference speed statistics.

On the proposed benchmark, we first evaluate \sparkvlnplain{} against mainstream VLN methods under both the Idealized and Realistic Dynamic Environments. All methods degrade once the simulator no longer pauses during inference, confirming observation staleness as a real threat. Besides, \sparkvlnplain{} attains the best success rate and social compliance in both stages. We further evaluate our fast planner alone on the point-goal social navigation~(PointNav) task under the same two stages, where it outperforms other learning-based planners. Ablations verify that token-wise streaming surpasses wait-then-act transfer by 10.0 points in SR, and runtime profiling shows that streaming cuts the per-update latency from 0.788\,s to 0.185\,s.

Our main contributions are as follows:
\begin{itemize}
    \item We propose \sparkvln{}, a dual-system framework for dynamic social VLN that streams hidden states in a token-wise manner from the slow VLM reasoner to the fast expert planner, so that the planner acts on fresh guidance throughout inference.
    \item We instantiate it with three coupling modules: a Token-Wise Hidden Streamer that \textit{extracts} intermediate hidden states during generation, a Sequence-to-Slot Latent Bridge that \textit{projects} them into fixed-size slots, and an Evolving Latent Conditioner that \textit{infuses} them into the flow-matching expert planner.   
    \item We introduce a human-centric benchmark suite for dynamic social VLN that keeps pedestrians moving throughout model inference, with goal-directed pedestrians and social-compliance metrics, on which \sparkvlnplain{} cuts latency and improves navigation performance compared with SOTA algorithms.
\end{itemize}
\section{Related Work}
\label{sec:related_work}

\begin{figure*}[t]
\centering
\includegraphics[width=0.75\linewidth]{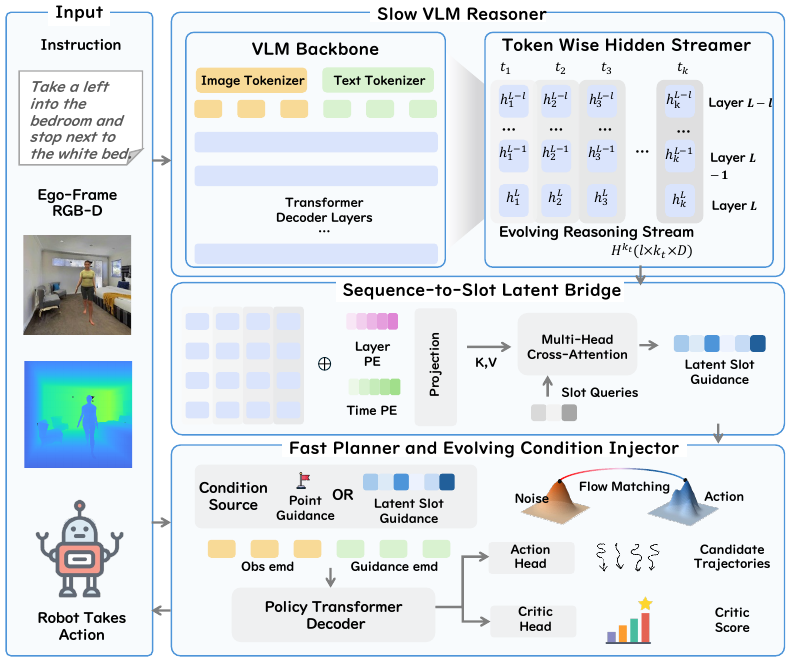}
\caption{\textbf{Overview of the \sparkvln{} framework.} \textbf{Top:} The VLM backbone processes egocentric RGB and instruction tokens; the \emph{Token-Wise Hidden Streamer} extracts hidden states from designated layers during autoregressive generation, producing an evolving stream $\mathbf{H}^{k_t} \in \mathbb{R}^{l \times k_t \times D}$. \textbf{Middle:} The \emph{Sequence-to-Slot Latent Bridge} augments the stream with layer and time positional encodings, projects it into the planner's embedding space, and compresses it into fixed-size latent slots via cross-attention with learnable queries. \textbf{Bottom:} The expert planner fuses observation and latent slot embeddings, generates candidate trajectories via rectified flow matching, and selects the action through a safety critic. As the VLM continues generating, the latent slots are re-computed and the planner re-plans.}

\label{fig:pipeline}
\end{figure*}

\subsection{Vision-Language Navigation}
\label{sec:rw-vln}
Vision-and-Language Navigation (VLN) has moved from discrete graph-based formulations~\cite{ krantz2020beyond,ku2020room,fried2018speaker,hong2021vln} to continuous, end-to-end policies driven by large pretrained backbones~\cite{bai2025qwen3,li2024llama,lin2024vila}. Early VLN systems such as NaVid~\cite{zhang2024navid}, MapNav~\cite{zhang2025mapnav}, and NaVILA~\cite{cheng2025navila} couple video-language pretraining with action decoding, while later systems tackle streaming history~\cite{zhang2024uni, wei2025streamvln, zeng2025janusvln}, and a parallel line scales toward general, cross-embodiment navigation models and unified training paradigms~\cite{zhang2025embodied, li2026gn0}, while others build explicit world models for visual navigation~\cite{dong2025towards, dong2026language} (see~\cite{kong2025survey} for a broader survey of 3D/4D world modeling) or pursue LLM-driven zero-shot exploration~\cite{gong2025stairway}. These gains, however, come from ever-larger VLMs, whose slow inference becomes a liability in dynamic scenes. The scene evolves during inference, so an action
selected by the slow VLM is outdated by the time it executes, undermining its
validity and safety.

\subsection{Real-Time and Latency-Aware VLAs}
\label{sec:rw-latency}
From the standpoint of system design and latency, prior work falls into three categories (see~\cite{hu2025avsurvey} for a broader survey of Vision-Language-Action (VLA) models). Early systems follow a \emph{reason-then-act} schedule~\cite{zhang2024navid, cheng2025navila, zhang2024uni,peng2026structured}: they emit an action chunk and then stall while the next is reasoned, so latency compounds into visible inter-chunk pauses. \emph{Real-time execution} frameworks instead overlap inference with execution through asynchronous pipelining, issuing the next inference before the current chunk is exhausted~\cite{black2026real,xie2026dynamicvla,hu2026ar,sendai2025leave}; this preserves motion continuity but not semantic freshness, since each chunk stays anchored to observations from the start of a long inference cycle. \emph{Dual-system} designs decouple a slow reasoning module from a fast action policy and run them at different frequencies, recovering reactivity while keeping large-model reasoning~\cite{zhu2024language, zhang2024hirt, han2024dpvla, li2024scvla, cui2025openhelix, lin2025onetwovla, song2025hume, chen2025fisvla, song2025rationalvla, huang2025thinkact, lu2026onevl, wei2026ground, wei2025streamvln, huang2026tic}; yet the two still communicate only through completed reasoning chains or finished action commands. Across all three, the unit passed from reasoning to action waits for a \emph{complete} output, a full action chunk or a single latent emitted only at the end of VLM inference. The policy therefore acts on stale guidance, blind to changes while the VLM reasons. \sparkvlnplain{} dissolves this inference blocking by refreshing the conditioning from the slow VLM to the fast planner at the \emph{token} level, so fresh guidance is available while the
VLM is reasoning rather than only at the end.

\subsection{Social and Dynamic Navigation Benchmarks}
\label{sec:rw-social}
Standard VLN benchmarks evaluate instruction following in static scenes: R2R and RxR~\cite{anderson2018vision, ku2020room} on discrete graphs and VLN-CE~\cite{krantz2020beyond} in continuous motion. A parallel line of work studies social navigation, focusing on goal reaching and pedestrian avoidance~\cite{biswas2022socnavbench, vuong2024habicrowd,gong2025cognition, hirose2023sacson, puig2023habitat, perezhigueras2023hunavsim}. While recent efforts attempt to bridge this gap by introducing dynamic human agents into language-guided tasks~\cite{dong2025ha, chen2025lisn, huang2026tic, peng2026freeaskworld}, existing platforms still fall short of a holistic evaluation. Existing benchmarks decouple linguistic reasoning from social compliance, evaluating the two in isolation. Even when dynamic agents are present in the VLN task, they lack temporal realism: the environment is frozen during inference, leaving latency risks and the lag between observation and execution unquantified. To address these limitations, we introduce an asynchronous benchmark suite for dynamic social VLN that unifies instruction-guided and social-compliant navigation under realistic inference-time dynamics. By keeping pedestrians moving throughout inference, it exposes inference latency and connects high-level language understanding with real-time, safety-critical planning.
\section{Problem Formulation}
\label{sec:method:problem}

\subsection{Vision-Language Navigation}
Given a natural-language instruction $\mathcal{I}$, an agent navigates toward a goal region $\mathcal{G}(\mathcal{I})$. At each discrete time step $t$, the agent receives an egocentric observation and pose:
\begin{equation}
    x_t = (o_t,\, p_t),
    \qquad
    o_t = (I_t^{\mathrm{rgb}},\, I_t^{\mathrm{depth}}).
\end{equation}
A policy $\pi$ maps the instruction and observation history to a navigation command $\mathbf{u}_t = \pi(\mathcal{I},\, x_{\leq t})$, inducing a rollout $\Gamma_\pi = \{(x_t, \mathbf{u}_t)\}_{t=0}^{T}$. The VLN objective is:
\begin{equation}
    \pi^\star
    =
    \arg\max_{\pi}\;
    \mathbb{E}\bigl[M(\Gamma_\pi,\,\mathcal{G}(\mathcal{I}))\bigr],
\end{equation}
where $M$ aggregates standard navigation metrics such as Success Rate and Success weighted by Path Length.

\subsection{Dual-System VLN}
A dual-system VLN policy consists of a \emph{slow} vision-language reasoner $f_{\mathrm{vlm}}$ and a \emph{fast} expert planner $f_{\mathrm{plan}}$:
\begin{equation}
     f_{\mathrm{vlm}}\!\bigl(\mathcal{I},\; x_{t'}\bigr)
    \mapsto
    \mathbf{z}_{t'},
    \quad
     \mathbf{c}_t = \phi(\mathbf{z}_{t'}),
    \quad
    f_{\mathrm{plan}}\!\bigl(x_t,\;\mathbf{c}_t\bigr)
    \mapsto
    \mathbf{u}_t ,
\end{equation}
where $t'$ denotes the time at which the VLM starts its inference, $\mathbf{z}_{t'}$ is the semantic reasoning state \textit{extracted} from the VLM, and $\phi$ \textit{projects} it into a planner-compatible conditioning signal $\mathbf{c}_t$ that is \textit{infused} into the expert planner. In conventional dual-system designs, the planner receives $\mathbf{c}_t$ only after the VLM completes inference from the earlier observation $x_{t'}$. It therefore uses a stale semantic signal with lag $t - t'$. In dynamic, human-centric environments where pedestrians move continuously, this gap is safety-critical: a maneuver deemed safe at $t'$ may already be unsafe at $t$.

\subsection{Streaming Knowledge Transfer}
\sparkvlnplain{} lets the slow VLM reasoner stream its knowledge to the fast expert planner throughout the VLM's generation. The VLM autoregressively produces intermediate hidden states
\begin{equation}
    \mathcal{H}_{t'}
    =
    \{\mathbf{h}_{t'}^{(1)},\, \mathbf{h}_{t'}^{(2)},\, \ldots,\,
    \mathbf{h}_{t'}^{(K)}\},
\end{equation}
where $K$ is the total token count and $\mathbf{h}_{t'}^{(k)}$ is the hidden state at the $k$-th token. Rather than waiting for the full output, $\phi_{\mathrm{stream}}$ \emph{extracts} hidden states from an intermediate layer in a token-wise manner, \emph{projects} the variable-length sequence into a fixed-dimensional representation, and \emph{infuses} it into the planner:
\begin{equation}
    \mathbf{c}_t = \phi_{\mathrm{stream}}\bigl(\mathbf{H}_{t'}^{(1:k_t)}\bigr),
    \qquad
    \mathbf{u}_t = f_{\mathrm{plan}}\bigl(x_t, \mathbf{c}_t\bigr),
\end{equation}
where $k_t$ is the latest token emitted by time $t$. This enables the planner to receive progressively refined semantic guidance from intermediate generated tokens, rather than remaining blind throughout the entire inference period.
\section{Method}
\label{sec:method}

\begin{figure*}[t]
\centering
\includegraphics[width=\linewidth]{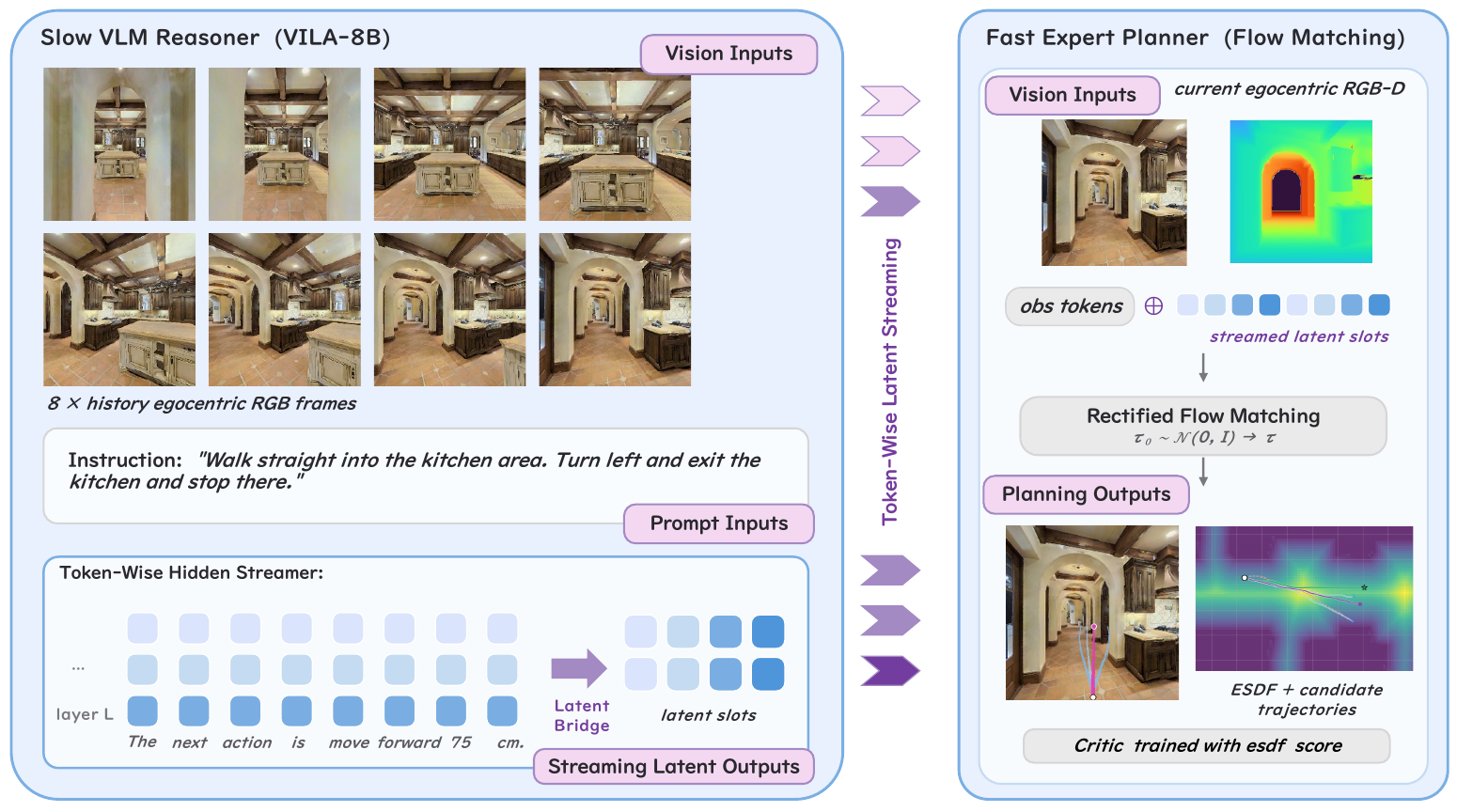}
\caption{\textbf{Data flow of \sparkvln{}.} The slow VLM reasoner receives the language instruction together with 8 recent egocentric RGB frames. As it autoregressively generates the answer (e.g., ``The next action is move forward 75 cm.''), the Token-Wise Hidden Streamer extracts intermediate-layer hidden states during generation, and the Sequence-to-Slot Latent Bridge compresses the evolving hidden stream into $N{=}8$ fixed-size latent slots that are streamed to the planner. The fast expert planner conditions on the current egocentric RGB-D observation fused with the streamed latent slots, and generates multiple candidate trajectories via rectified flow matching. A safety critic supervised by the Euclidean Signed Distance Field (ESDF) scores the candidates and selects the executed trajectory.}
\label{fig:vln_overview}
\end{figure*}

\sparkvlnplain{}, as shown in Fig.~\ref{fig:pipeline}, is a dual-system framework in which a slow VLM reasoner and a fast expert planner run concurrently, connected by a streaming knowledge transfer pipeline that delivers guidance in a token-wise manner. At each timestep, the robot receives an egocentric RGB-D observation and a natural-language instruction. The VLM reasoner (VILA-8B~\cite{lin2024vila}) processes the visual input together with the instruction and begins autoregressive generation. The \textbf{Token-Wise Hidden Streamer} (\ref{sec:method:streamer}) \emph{extracts} intermediate hidden states $\mathbf{h}_{\mathcal{L}}^{(k)}$ during generation, producing a growing prefix of latent representations. The \textbf{Sequence-to-Slot Latent Bridge} (\ref{sec:method:bridge}) \emph{projects} this variable-length, high-dimensional prefix into a fixed set of compact latent slots that lie in the expert planner's conditioning space. The \textbf{Evolving Latent Conditioner} (\ref{sec:method:conditioner}) \emph{infuses} these slots into the rectified-flow-matching expert planner, which generates a continuous trajectory. As the VLM continues generating, the prefix grows, the slots are re-computed, and the planner re-plans.

\subsection{Token-Wise Hidden Streamer}
\label{sec:method:streamer}
Given the multimodal input $\mathbf{x}_{t'}$, the VLM backbone~\cite{lin2024vila} autoregressively generates a sequence of $K$ tokens. At each decoding step $k \in \{1, \dots, K\}$, the hidden state of the $l$-th transformer layer is updated based on the historical context as:
\begin{equation}
  \mathbf{h}_{l}^{(k)}
  = \mathbf{F}\bigl(\mathbf{x}_{t'} \mid l, k, \theta\bigr)
  \;\in\; \mathbb{R}^{D},
  \label{eq:vlm_forward}
\end{equation}
where $D$ denotes the VLM hidden dimension, $\theta$ denotes the frozen VLM parameters, and $\mathbf{F}(\cdot \mid l, k, \theta)$ extracts the feature of the $l$-th layer at decoding step $k$. Standard frameworks block downstream execution until the entire sequence of $K$ tokens is generated, leading to information latency. To circumvent this bottleneck, the Token-Wise Hidden Streamer $\phi_{\mathrm{stream}}$ extracts these internal representations on the fly. Specifically, we define a designated subset of decoder layers $\mathcal{L} = \{l_1, \dots, l_L\}$. For the token generated at step $k$, the streamer collects its hidden states across all layers in
$\mathcal{L}$ to form a per-token multi-layer representation:
\begin{equation}
  \mathbf{h}_{\mathcal{L}}^{(k)}
  = \bigl[\mathbf{h}_{l_1}^{(k)};\, \mathbf{h}_{l_2}^{(k)};\, \dots;\,
  \mathbf{h}_{l_L}^{(k)}\bigr]
  \;\in\; \mathbb{R}^{L \times D},
  \label{eq:spatial_snapshot}
\end{equation}

As decoding progresses up to time step $t$, the VLM has emitted $k_t$ tokens ($k_t \le K$). The cumulative hidden stream available to the downstream architecture is formalized as:
\begin{equation}
  \mathbf{H}_{t'}^{(k_t)}
  = \bigl[\mathbf{h}_{\mathcal{L}}^{(1)},\, \mathbf{h}_{\mathcal{L}}^{(2)},\,
  \dots,\, \mathbf{h}_{\mathcal{L}}^{(k_t)}\bigr]
  \;\in\; \mathbb{R}^{L \times k_t \times D}.
  \label{eq:hidden_stream_tensor}
\end{equation}
By aggregating these per-token multi-layer representations step by step, $\mathbf{H}_{t'}^{(k_t)}$ provides immediate, progressively refined semantic guidance for the expert planner throughout the VLM's inference process, mitigating observation staleness before generation completes.

\subsection{Sequence-to-Slot Latent Bridge}
\label{sec:method:bridge}
The cumulative hidden stream $\mathbf{H}_{t'}^{(k_t)}$ possesses a dynamically expanding temporal dimension $k_t$ and resides in the high-dimensional reasoning space of the VLM. To seamlessly condition the downstream expert planner, which requires a fixed-size input and operates on a lower dimension, $\phi_{\mathrm{stream}}$ introduces the Sequence-to-Slot Latent Bridge. This module serves as a cross-modal translator, distilling the variable-length, multi-layer features into a compact, fixed-dimensional set of latent tokens. Because the raw hidden states extracted from the streamer lack explicit structural markers regarding their layer origin and generation step, we first augment the tensor with factorized spatio-temporal positional
encodings:
\begin{equation}
  \mathbf{m}_{l,k} = \mathbf{h}_{l}^{(k)} + \mathbf{e}_{\mathrm{layer}}(l)
  + \mathbf{e}_{\mathrm{time}}(k),
  \label{eq:pe}
\end{equation}
where $\mathbf{e}_{\mathrm{layer}}(l), \mathbf{e}_{\mathrm{time}}(k) \in \mathbb{R}^{D}$ are learnable positional encodings for the layer index and the decoding step, respectively. We collect and flatten the enriched tokens into a unified memory tensor $\mathbf{M}_{t'}^{(k_t)} \in \mathbb{R}^{(L \cdot k_t) \times D}$. To bridge the cross-modal manifold gap and squeeze this memory into a fixed-capacity representation, we instantiate $N$ learnable query vectors $\mathbf{Q} = [\mathbf{q}_1, \dots, \mathbf{q}_N]^\top \in \mathbb{R}^{N \times d}$ to define our latent slots, following a Perceiver-style cross-attention bottleneck~\cite{jaegle2021perceiver}, where $d$ is the planner's input dimension. We first project the high-dimensional enriched memory into the low-dimensional planning space via:
\begin{equation}
  \widetilde{\mathbf{M}}_{t'}^{(k_t)}
  = \mathbf{M}_{t'}^{(k_t)} \mathbf{W}_p
  \;\in\; \mathbb{R}^{(L \cdot k_t) \times d},
  \label{eq:projection}
\end{equation}
where $\mathbf{W}_p \in \mathbb{R}^{D \times d}$ is a learnable projection matrix. The latent slots then adaptively harvest relevant contextual abstractions across the spatio-temporal memory via multi-head cross-attention. The update for the $i$-th latent slot is formulated as:
\begin{equation}
  \mathbf{z}_i = \mathbf{q}_i
  + \operatorname{CrossAttn}\bigl(\mathbf{q}_i,\;
  \widetilde{\mathbf{M}}_{t'}^{(k_t)}\bigr),
  \label{eq:cross_attn}
\end{equation}
followed by a standard feed-forward network to yield the finalized slots. We stack these slots to compose the fixed-dimensional conditioning
matrix:
\begin{equation}
  \mathbf{R}_t = \bigl[\mathbf{z}_1, \mathbf{z}_2, \dots,
  \mathbf{z}_N\bigr]^\top \;\in\; \mathbb{R}^{N \times d}.
  \label{eq:roller}
\end{equation}
Unlike naive pooling operations that collapse structural token-level granularity, this cross-attention bridge dynamically routes the most informative spatio-temporal features into a fixed, expressive communication channel tailored for real-time planner infusion.

\subsection{Evolving Latent Conditioner and Expert Planner}
\label{sec:method:conditioner}
We model reactive navigation as sampling a future trajectory
$\tau \in \mathbb{R}^{H \times 3}$ of $H$ waypoints from a conditional
generator:
\begin{equation}
  \tau_1 = G_\psi\!\bigl(\tau_0,\; \mathbf{c}_t\bigr),
  \qquad
  \tau_0 \sim \mathcal{N}(0, \mathbf{I}),
  \label{eq:cond_gen}
\end{equation}
where $\tau_1$ denotes the generated trajectory, and the noise realization $\tau_0$ preserves the multi-modality of viable paths in dense social spaces. This expert planner is driven by a unified conditioning matrix $\mathbf{c}_t = \bigl[\, \mathbf{z}_{\mathrm{obs}};\; \mathbf{z}_{\mathrm{g}}^{(m)} \,\bigr]$, which pairs egocentric spatial tokens $\mathbf{z}_{\mathrm{obs}}$ with a flexible goal token $\mathbf{z}_{\mathrm{g}}^{(m)}$ across task modalities $m$:
\begin{equation}
  \mathbf{z}_{\mathrm{g}}^{(m)}
  =
  \begin{cases}
    \varnothing, & m = \text{no-goal},\\[2pt]
    \mathbf{W}_g\, g, & m = \text{point-goal},\\[2pt]
    \mathbf{R}_t, & m = \text{language-goal}.
  \end{cases}
  \label{eq:full_condition_assembly}
\end{equation}

This unified formulation enables the planner to learn foundational navigation policies and obstacle avoidance from abundant basic navigation data and seamlessly transfer to the VLN task. Under the language-goal setting, the streamed guidance slots $\mathbf{R}_t$ are dynamically re-computed as generation proceeds, allowing the conditioning context to evolve smoothly within a single VLM reasoning pass. We parameterize the trajectory generator $G_\psi$ via conditional rectified flow matching~\cite{liu2023flow,lipman2023flow}, following recent flow-based VLA policies such as $\pi_0$~\cite{black2024pi0} rather than a diffusion-based formulation~\cite{chi2023diffusion}, defining a linear interpolant between noise and the expert trajectory over flow time $s \in [0,1]$:
\begin{equation}
  \tau_s = (1-s)\,\tau_0 + s\,\tau_1,
  \label{eq:interpolant}
\end{equation}
whose theoretical target velocity $\tau_1 - \tau_0$ remains constant across $s$. A transformer-based vector field network $v_\psi$ is optimized to regress this velocity by minimizing the empirical rectifying loss:
\begin{equation}
  \mathcal{L}_{\mathrm{RF}}
  =
  \mathbb{E}_{s \sim \mathcal{U}[0,1],\, \tau_0,\, \tau_1}
  \Bigl[\,\bigl\| v_\psi(s,\, \tau_s,\, \mathbf{c}_t)
  - (\tau_1 - \tau_0) \bigr\|^2 \,\Bigr].
  \label{eq:rf_loss}
\end{equation}
The straight transport path admits accurate few-step integration. At deployment, each candidate is decoded by $J$-step Euler integration with the latest $\mathbf{c}_t$:
\begin{equation}
  \tau^{(j+1)} = \tau^{(j)} + \tfrac{1}{J}\,
  v_\psi\!\left(\tfrac{j}{J},\, \tau^{(j)},\, \mathbf{c}_t\right),
  \label{eq:euler}
\end{equation}
where the superscript $(j)$ indexes the integration step. Sampling multiple noise realizations yields a set of $N_c$ candidate trajectories $\{\tau^{[n]}\}_{n=1}^{N_c}$, from which a goal-agnostic safety critic selects the executed one:
\begin{equation}
  \tau^* = \arg\max_{\tau^{[n]}}\;
  C_\omega\!\bigl(\tau^{[n]},\, \mathbf{z}_{\mathrm{obs}}\bigr),
  \label{eq:critic}
\end{equation}
which evaluates only collision-freeness and kinematic smoothness from $\mathbf{z}_{\mathrm{obs}}$, isolating reactive safety filtering from high-level intent generation.
\section{Experiments}
\label{sec:exps}

\begin{figure*}[t]
\centering
\includegraphics[width=\textwidth]{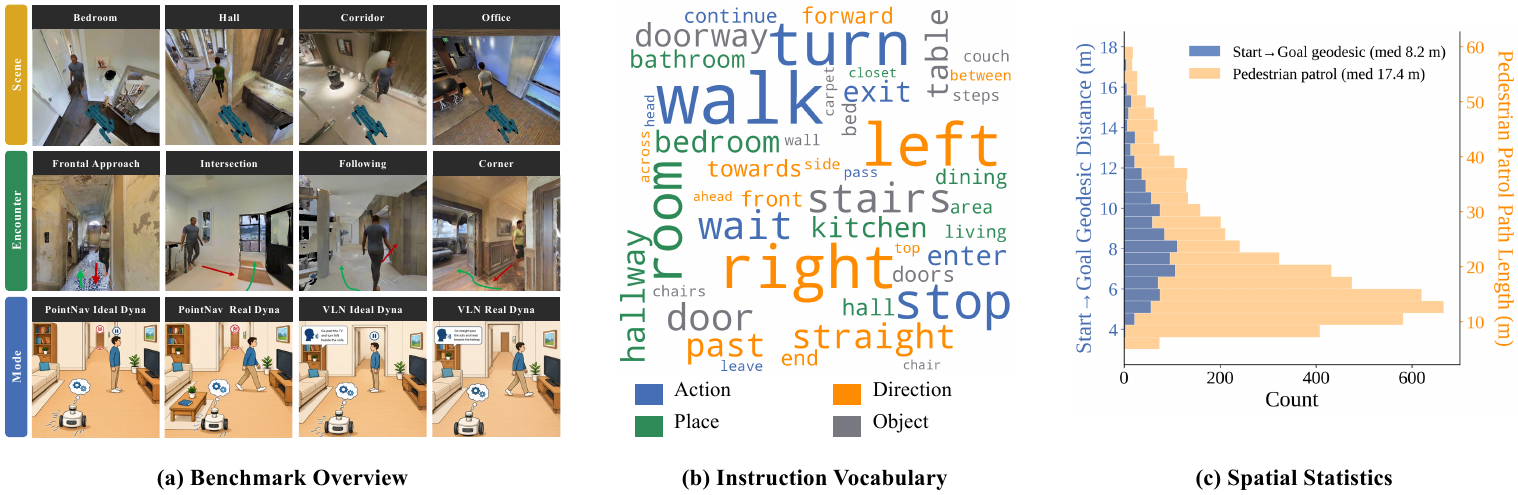}
\caption{\textbf{Overview of our dynamic social VLN benchmark.}
\textbf{(a) Benchmark Overview.} The benchmark spans diverse indoor \emph{scenes}
(such as bedroom, office, hall, and corridor) and covers several human \emph{encounter} patterns, including frontal approach, intersection, following, and corner, with the robot path in \textcolor{sparkvln_green}{green} and the pedestrian path in \textcolor{sparkvln_red}{red}. It supports both the PointNav and VLN tasks under two evaluation \emph{modes}, the Idealized Dynamic Environment (with pedestrians, blocking simulation) and the Realistic Dynamic Environment (with pedestrians, non-blocking simulation). \textbf{(b) Instruction Vocabulary.} Word cloud of navigation instructions, colored by semantic role, namely \textcolor{sparkvln_blue}{action}, \textcolor{sparkvln_orange}{direction}, \textcolor{sparkvln_green}{place}, and \textcolor{sparkvln_gray}{object}. \textbf{(c) Spatial Statistics.} Distributions of start-to-goal geodesic distance (\textcolor{sparkvln_blue}{blue}, median $8.2$~m) and per-pedestrian patrol path length (\textcolor{sparkvln_orange}{orange}, median $17.4$~m), showing that pedestrian trajectories are long enough to repeatedly intersect the agent's route.}
\label{fig:benchmark_overview}
\end{figure*}

\begin{figure*}[t]
\centering
\includegraphics[width=\textwidth]{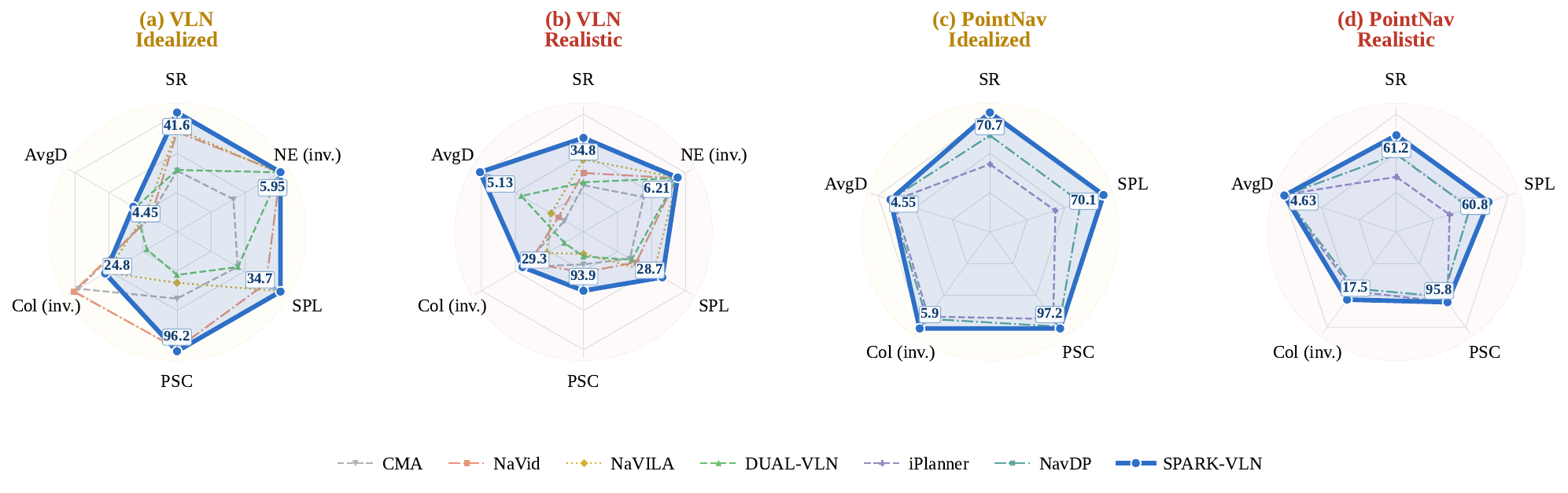}
\caption{\textbf{Radar comparison across tasks and dynamic environments.}
\textbf{(a,b)} Vision-Language Navigation (VLN) and \textbf{(c,d)} point-goal navigation, each evaluated under the Idealized and Realistic dynamic environments.
\sparkvln{} achieves the largest coverage in each setting, with annotated raw values. The performance gap between the Idealized and Realistic panels, i.e., (a)~vs.~(b) and (c)~vs.~(d), reflects each method's robustness to observation staleness induced by inference latency.}
\label{fig:radar_4panel}
\vspace{-2em}
\end{figure*}

We first describe the evaluation setup (\ref{sec:exp_setup}), then present quantitative comparisons on vision-language navigation and point-goal social navigation (\ref{sec:exp_quant}). A series of ablation studies validates the contributions of the latent-extraction strategy, the bridge design, and the runtime behavior (\ref{sec:exp_ablation}).

\subsection{Experimental Setup}
\label{sec:exp_setup}

\begin{table*}[t]
\centering
\caption{Evaluation of \textbf{VLN methods} on the dynamic social VLN benchmark.
\textit{Idealized Dynamic Environment}: pedestrians move but the simulator pauses during inference;
\textit{Realistic Dynamic Environment}: the simulator runs continuously during inference, so pedestrians advance during model inference.
$\uparrow$ higher is better; $\downarrow$ lower is better. \textbf{Bold} denotes best; \underline{underline} denotes second-best within each mode.}
\label{tab:vln_3stage}
\resizebox{\textwidth}{!}{
\begin{tabular}{l|c|l|c|ccc|ccc}
\toprule
\multirow{2}{*}{\textbf{Method}} 
& \multirow{2}{*}{\textbf{Venue}}
& \multirow{2}{*}{\textbf{MLLM}}
& \multirow{2}{*}{\textbf{Params}}
& \multicolumn{3}{c|}{\textbf{Task Completion}} 
& \multicolumn{3}{c}{\textbf{Social Interaction}} \\
\cmidrule(lr){5-7} \cmidrule(lr){8-10}
& & & 
& SR$\uparrow$ & NE$\downarrow$ & SPL$\uparrow$
& PSC$\uparrow$ & Col$\downarrow$ & AvgD$\uparrow$ \\
\midrule
\rowcolor{sparkvln_yellow!15}
\multicolumn{10}{l}{\textcolor{sparkvln_yellow}{\ding{108}}~\textbf{Idealized Dynamic Environment (With Pedestrians, Blocking Simulation)}} \\
\midrule
\rowcolor{gray!15}
\multicolumn{10}{l}{\textit{\textbf{VLN (Single-system)}}} \\
Seq2Seq~\cite{krantz2020beyond} 
& ECCV'20 & -- & -- 
& 17.60 & 8.32 & 16.61 & 94.4 & 17.3 & 4.28 \\
CMA~\cite{krantz2020beyond}     
& ECCV'20 & -- & -- 
& 26.00 & 7.27 & 24.25 & 94.2 & \underline{13.8} & 4.26 \\
NaVid~\cite{zhang2024navid}     
& RSS'24 
& \raisebox{-0.5ex}{\includegraphics[width=0.018\linewidth]{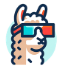}}~LLaMA-VID~\cite{li2024llama} 
& 7B
& 36.40 & \underline{5.96} & 31.19 & \underline{96.1} & \textbf{11.8} & 4.25 \\
Uni-NaVid~\cite{zhang2024uni}   
& RSS'25 
& \raisebox{-0.5ex}{\includegraphics[width=0.018\linewidth]{figs/icons/llama-vid.png}}~LLaMA-VID~\cite{li2024llama} 
& 7B
& 36.00 & 5.98 & 31.97 & 94.5 & 21.6 & 4.15 \\
NaVILA~\cite{cheng2025navila}   
& RSS'25 
& \raisebox{-0.5ex}{\includegraphics[width=0.018\linewidth]{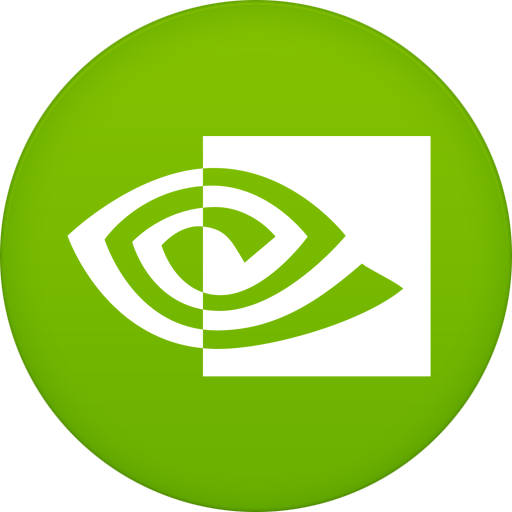}}~VILA~\cite{lin2024vila} 
& 8B
& \underline{37.20} & 5.99 & \underline{34.47} & 93.6 & 25.4 & 4.31 \\
\rowcolor{gray!15}
\multicolumn{10}{l}{\textit{\textbf{VLN (Dual-system)}}} \\
DualVLN~\cite{wei2026ground}   
& ICLR'26 
& \raisebox{-0.5ex}{\includegraphics[width=0.018\linewidth]{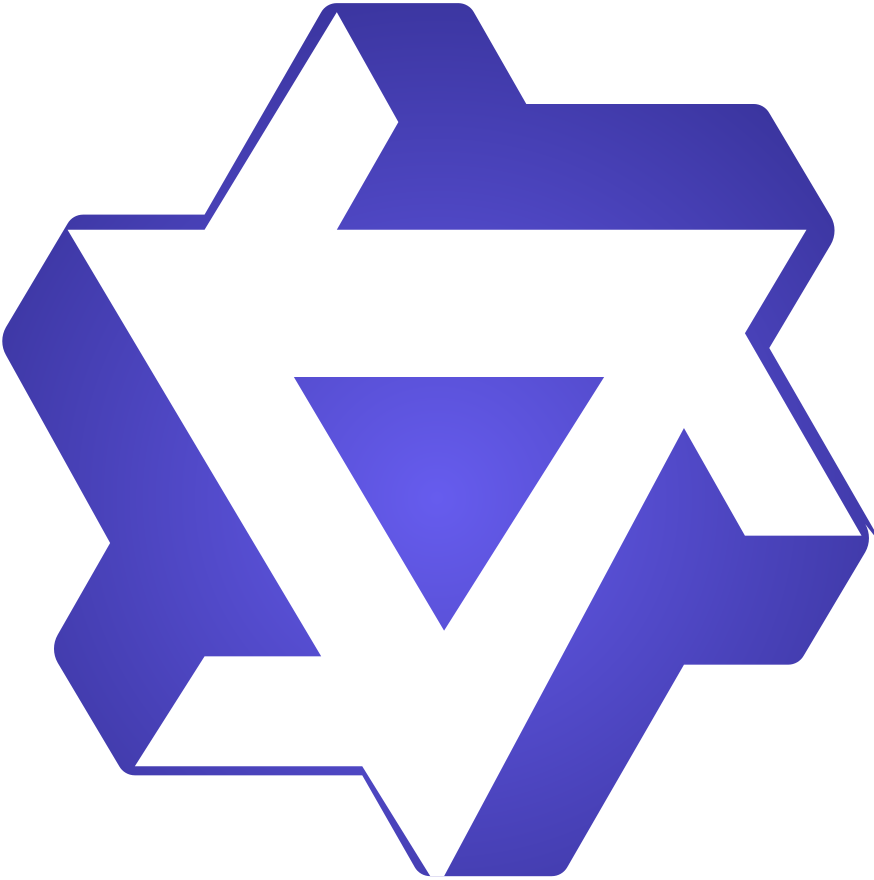}}~Qwen-VL-2.5~\cite{Qwen2.5-VL} 
& 7B
& 26.20 & \underline{5.96} & 24.33 & 93.3 & 42.0 & \underline{4.43} \\
\rowcolor{sparkvln_blue!8}
\textbf{\sparkvln}                   
& -- 
&  \raisebox{-0.5ex}{\includegraphics[width=0.018\linewidth]{figs/icons/vila.png}}~VILA~\cite{lin2024vila}
& 8B
& \textbf{41.60} & \textbf{5.95} & \textbf{34.67} & \textbf{96.2} & 24.8 & \textbf{4.45} \\
\midrule

\rowcolor{sparkvln_red!15}
\multicolumn{10}{l}{\textcolor{sparkvln_red}{\ding{108}}~\textbf{Realistic Dynamic Environment (With Pedestrians, Non-Blocking Simulation)}} \\
\midrule
\rowcolor{gray!15}
\multicolumn{10}{l}{\textit{\textbf{VLN (Single-system)}}} \\
Seq2Seq~\cite{krantz2020beyond} 
& ECCV'20 & -- & -- 
& 15.50 & 8.31 & 14.16 & 93.0 & \underline{29.4} & 4.24 \\
CMA~\cite{krantz2020beyond}     
& ECCV'20 & -- & -- 
& 22.10 & 7.16 & 20.81 & 92.9 & \underline{29.4} & 4.17 \\
NaVid~\cite{zhang2024navid}     
& RSS'24 
& \raisebox{-0.5ex}{\includegraphics[width=0.018\linewidth]{figs/icons/llama-vid.png}}~LLaMA-VID~\cite{li2024llama} 
& 7B
& 25.40 & 6.24 & 22.35 & \underline{93.2} & 32.1 & 4.24 \\
Uni-NaVid~\cite{zhang2024uni}   
& RSS'25 
& \raisebox{-0.5ex}{\includegraphics[width=0.018\linewidth]{figs/icons/llama-vid.png}}~LLaMA-VID~\cite{li2024llama} 
& 7B
& \underline{30.20} & 6.23 & \underline{26.93} & 92.9 & 34.6 & 4.10 \\
NaVILA~\cite{cheng2025navila}   
& RSS'25 
& \raisebox{-0.5ex}{\includegraphics[width=0.018\linewidth]{figs/icons/vila.png}}~VILA~\cite{lin2024vila} 
& 8B
& 29.00 & \underline{6.22} & 26.63 & 92.5 & 39.3 & 4.32 \\
\rowcolor{gray!15}
\multicolumn{10}{l}{\textit{\textbf{VLN (Dual-system)}}} \\
DualVLN~\cite{wei2026ground}   
& ICLR'26 
& \raisebox{-0.5ex}{\includegraphics[width=0.018\linewidth]{figs/icons/qwen.png}}~Qwen-VL-2.5~\cite{Qwen2.5-VL} 
& 7B
& 22.90 & 6.23 & 21.23 & 92.6 & 46.6 & \underline{4.66} \\
\rowcolor{sparkvln_blue!8}
\textbf{\sparkvln}                   
& -- 
&  \raisebox{-0.5ex}{\includegraphics[width=0.018\linewidth]{figs/icons/vila.png}}~VILA~\cite{lin2024vila}
& 8B
& \textbf{34.80} & \textbf{6.21} & \textbf{28.68} & \textbf{93.9} & \textbf{29.3} & \textbf{5.13} \\

\bottomrule
\end{tabular}
}
\vspace{-2mm}
\end{table*}
\begin{table}[t]
\centering
\caption{Evaluation of \textbf{point-goal social navigation} on the dynamic social benchmark.
\textit{Idealized Dynamic Environment}: the simulator pauses during inference, so the agent acts on fresh observations;
\textit{Realistic Dynamic Environment}: the simulator runs continuously, so pedestrians advance and observations become stale at action time.
Methods in \textcolor{gray}{gray} use a privileged environment mesh and mesh-derived reference path (oracle, not direct competitors).
$\uparrow$ higher is better; $\downarrow$ lower is better. \textbf{Bold} best; \underline{underline} second-best (excluding oracles).}
\label{tab:pointgoal_3stage}
\resizebox{\columnwidth}{!}{%
\begin{tabular}{l|cc|ccc}
\toprule
\multirow{2}{*}{\textbf{Method}}
& \multicolumn{2}{c|}{\textbf{Task Completion}}
& \multicolumn{3}{c}{\textbf{Social Interaction}} \\
\cmidrule(lr){2-3} \cmidrule(lr){4-6}
& SR$\uparrow$ & SPL$\uparrow$
& PSC$\uparrow$ & Col$\downarrow$ & AvgD$\uparrow$ \\
\midrule
\rowcolor{sparkvln_yellow!15}
\multicolumn{6}{l}{\textcolor{sparkvln_yellow}{\ding{108}}~\textbf{Idealized Dynamic Environment}} \\
\midrule
\rowcolor{gray!15}
\multicolumn{6}{l}{\textit{\textbf{Classical / Rule-based (Oracle)}}} \\
\textcolor{gray}{PathFollower}                          & \textcolor{gray}{60.0} & \textcolor{gray}{55.9} & \textcolor{gray}{93.6} & \textcolor{gray}{33.9} & \textcolor{gray}{4.05} \\
\textcolor{gray}{ORCA~\cite{vanDenBerg2011ORCA}}         & \textcolor{gray}{33.2} & \textcolor{gray}{32.9} & \textcolor{gray}{96.2} & \textcolor{gray}{16.4} & \textcolor{gray}{4.08} \\
\rowcolor{gray!15}
\multicolumn{6}{l}{\textit{\textbf{Learning-based}}} \\
iPlanner~\cite{yang2023iplanner}      & 49.1 & 49.1 & 96.7 & 10.7 & 4.53 \\
NavDP~\cite{cai2025navdp}             & \underline{61.1} & \underline{60.2} & \underline{97.1} & \underline{9.8} & \underline{4.54} \\
\rowcolor{sparkvln_blue!8}
\textbf{Ours-S1}                    & \textbf{70.7} & \textbf{70.1} & \textbf{97.2} & \textbf{5.9} & \textbf{4.55} \\
\midrule
\rowcolor{sparkvln_red!15}
\multicolumn{6}{l}{\textcolor{sparkvln_red}{\ding{108}}~\textbf{Realistic Dynamic Environment}} \\
\midrule
\rowcolor{gray!15}
\multicolumn{6}{l}{\textit{\textbf{Classical / Rule-based (Oracle)}}} \\
\textcolor{gray}{PathFollower}                          & \textcolor{gray}{56.7} & \textcolor{gray}{52.8} & \textcolor{gray}{93.0} & \textcolor{gray}{37.6} & \textcolor{gray}{4.05} \\
\textcolor{gray}{ORCA~\cite{vanDenBerg2011ORCA}}         & \textcolor{gray}{30.0} & \textcolor{gray}{29.7} & \textcolor{gray}{95.4} & \textcolor{gray}{21.5} & \textcolor{gray}{4.11} \\
\rowcolor{gray!15}
\multicolumn{6}{l}{\textit{\textbf{Learning-based}}} \\
iPlanner~\cite{yang2023iplanner}      & 43.8 & 43.8 & \underline{95.6} & \underline{20.9} & \textbf{4.64} \\
NavDP~\cite{cai2025navdp}             & \underline{53.2} & \underline{52.6} & 95.5 & 21.9 & 4.60 \\
\rowcolor{sparkvln_blue!8}
\textbf{Ours-S1}                    & \textbf{61.2} & \textbf{60.8} & \textbf{95.8} & \textbf{17.5} & \underline{4.63} \\
\bottomrule
\end{tabular}%
}
\vspace{-2mm}
\end{table}

\noindent\textbf{Dynamic Social VLN Benchmark.}
Most VLN benchmarks run synchronous evaluation loops that pause the simulator during model inference, freezing all pedestrians and suppressing observation staleness entirely, which hides whether inference latency changes navigation outcomes at all. To measure this effect directly, we introduce a benchmark suite with two evaluation stages.
\begin{itemize}
  \item \textbf{Idealized Dynamic Environment.} Pedestrians are present and move according to ORCA~\cite{vanDenBerg2011ORCA}, but the simulator pauses during each inference step. The agent therefore always acts on fresh observations, and this stage serves as the latency-free setting.
    \item \textbf{Realistic Dynamic Environment.} The simulator runs continuously, so pedestrians advance while the model computes and inference delay affects the results. Contrasting this stage against the idealized setting quantifies how well a method handles observation staleness gap.
\end{itemize}

As illustrated in Fig.~\ref{fig:benchmark_overview}, the benchmark spans diverse indoor scenes and covers both PointNav and VLN tasks under the Idealized and Realistic Dynamic Environments. For VLN tasks, the accompanying instruction vocabulary covers action, direction, place, and object roles, and the spatial statistics report a median pedestrian patrol path of $17.4$~m against an $8.2$~m median agent start-to-goal distance. Pedestrian density is scaled to scene area to avoid artificial overcrowding, and the scenes contain diverse social interaction patterns, such as frontal approach, intersection, following, and corner. Goal-directed pedestrians avoid one another via ORCA~\cite{vanDenBerg2011ORCA}.

\noindent\textbf{Metrics.}
In contrast to static VLN evaluation, where a single optimal reference path can be precomputed, our dynamic social benchmark features continuously moving pedestrians that reshape traversable space throughout inference. We therefore center evaluation on task completion and social interaction quality, adopting community guidelines for VLN evaluation~\cite{anderson2018vision,anderson2018evaluation} and social navigation evaluation~\cite{francis2023principles}. Task-completion metrics comprise Success Rate (SR), Navigation Error (NE), and Success weighted by Path Length (SPL). Social-interaction metrics comprise Personal Space Compliance (PSC), Human Collision rate (Col), and Average Distance to the nearest pedestrian (AvgD).

\noindent\textbf{Baselines.}
For VLN, we compare against five single-system methods and one dual-system baseline. Seq2Seq~\cite{krantz2020beyond} is a recurrent instruction-to-action policy, and CMA~\cite{krantz2020beyond} augments it with cross-modal attention between language and vision. NaVid~\cite{zhang2024navid} is a video-based VLM that predicts the next action from egocentric history, Uni-NaVid~\cite{zhang2024uni} extends it into a unified model across embodied navigation tasks, and NaVILA~\cite{cheng2025navila} couples a VILA backbone with an action decoder for legged navigation. The dual-system baseline DualVLN~\cite{wei2026ground} pairs a slow reasoner with a fast policy but hands over guidance only after the reasoner finishes. For PointNav, we compare against two classical methods with oracle access to the environment mesh and mesh-derived reference paths, namely PathFollower, a waypoint tracker that follows a precomputed reference path, and ORCA~\cite{vanDenBerg2011ORCA}, a reciprocal velocity-based collision-avoidance controller, as well as two learning-based planners, iPlanner~\cite{yang2023iplanner}, an imperative path planner, and NavDP~\cite{cai2025navdp}, a diffusion-based navigation policy trained with privileged guidance.

\noindent\textbf{Implementation Details.}
The VLM reasoner uses a VILA~\cite{lin2024vila} backbone with 8B parameters. The Token-Wise Hidden Streamer extracts hidden states from a designated subset of decoder layers at every generated token (layers 16--23). The Sequence-to-Slot Latent Bridge uses $N{=}8$ learnable queries with multi-head cross-attention to produce fixed-size slots. The expert planner is a rectified flow-matching network that generates continuous trajectories with 5-step Euler integration, followed by a goal-agnostic safety critic for trajectory selection. The expert planner is trained on point-goal navigation and natively supports the PointNav task. To adapt it to VLN, we add the streamed latent slot guidance as an additional conditioning input, so at VLN evaluation time the planner is given no goal coordinate~(set to zero) and is driven purely by the language instruction through the streamed slots. Fig.~\ref{fig:vln_overview} illustrates the full VLN pipeline of \sparkvlnplain{}.

\subsection{Quantitative Results}
\label{sec:exp_quant}

\noindent\textbf{Vision-language navigation.}
Tab.~\ref{tab:vln_3stage} reports our main results on the dynamic social VLN benchmark. In the Idealized Dynamic Environment, our model achieves the highest SR $41.60\%$ and SPL $34.67\%$ among all methods, while also attaining the best PSC $96.2\%$ and the lowest NE $5.95$~m. Compared to the strongest single-system baseline NaVILA, \sparkvlnplain{} improves SR by $4.4$ points. Compared to the dual-system baseline DualVLN, the gap widens to $15.4$ points, demonstrating the benefit of streaming knowledge transfer over the paradigm in which the fast planner receives guidance only after VLM reasoning completes.

Moving from the Idealized to the Realistic Dynamic Environment, every method degrades once observations go stale, confirming that observation staleness is a real problem in dynamic social navigation. Tab.~\ref{tab:runtime_profile} makes the mechanism explicit, and the heavier reasoners suffer most, with video-based VLMs such as NaVid losing close to a third of their SR. By contrast, our model relies on token-wise latent guidance to keep its effective latency low, and is among the most robust, shedding only about one-sixth of its SR while remaining the top performer in both stages. Crucially, conventional VLN benchmarks miss this entirely, since their blocking protocol freezes the environment during inference. Our asynchronous benchmark exposes this gap and is therefore essential for faithfully evaluating navigation in dynamic social environments.

In the Realistic Dynamic Environment, the advantage of \sparkvlnplain{} becomes more pronounced. Our model achieves $34.80\%$ SR and $28.68\%$ SPL, outperforming NaVILA ($29.00\%$ SR, $26.63\%$ SPL) and DualVLN ($22.90\%$ SR, $21.23\%$ SPL). Crucially, \sparkvlnplain{} also leads on the social metrics, attaining the best PSC $93.9\%$, the lowest collision rate $29.3\%$, and the highest AvgD $5.13$~m. These results confirm that streaming intermediate reasoning latents to the planner mitigates the observation staleness that degrades competing methods under realistic conditions.

\noindent\textbf{Point-goal social navigation.}
We also evaluate the PointNav-pretrained fast planner~(Ours-S1) in isolation on the point-goal task. As shown in Tab.~\ref{tab:pointgoal_3stage}, Ours-S1 achieves the highest SR $70.7\%$ and SPL $70.1\%$ among all learning-based methods in the Idealized Dynamic Environment, surpassing NavDP by $9.6$ points in SR and $9.9$ points in SPL. Social compliance is also the best, with the
collision rate dropping to $5.9\%$, nearly half that of NavDP ($9.8\%$).

In the Realistic Dynamic Environment, all methods degrade, yet our planner retains a clear lead, reaching $61.2\%$ SR and $60.8\%$ SPL versus $53.2\%$ and $52.6\%$ for NavDP. Notably, the oracle PathFollower, which has privileged access to the environment mesh, drops from $60.0\%$ to $56.7\%$ SR, while our learning-based planner exceeds it in both environments. This confirms that the flow-matching expert planner provides strong navigation and robust social compliance on its own, giving \sparkvlnplain{} a solid foundation for the streamed language guidance to build on.

\subsection{Ablation Studies}
\label{sec:exp_ablation}
We conduct ablation studies to validate two design decisions. First, whether streaming per-token hidden states improves over waiting for the full VLM output. Second, how to compress the streamed tokens into fixed-size planner conditioning, for which we compare a mean-pooling and a cross-attention bridge. All ablations are run on the Realistic Dynamic Environment, the setting in which latency actually matters.

\noindent\textbf{Knowledge-transfer strategy.}
We compare two knowledge-transfer strategies between the slow reasoner and the fast planner. \emph{Wait-then-Act} (W-t-A) blocks the planner until the VLM produces its full output and then infuses a single conditioning signal once per slow cycle, mirroring the conventional dual-system design. \emph{Stream} (Ours) instead updates the conditioning via the Token-Wise Hidden Streamer and the Sequence-to-Slot Latent Bridge, letting the planner act on progressively refined guidance while the VLM is still reasoning. As shown in Tab.~\ref{tab:ablation_streaming}, Stream improves SR from $24.80\%$ to $34.80\%$ ($+10.00$ points) and SPL from $21.64\%$ to $28.68\%$ ($+7.04$ points) over W-t-A, while reducing NE from $7.45$~m to $6.21$~m. The social metrics improve consistently, with PSC rising to $93.9\%$, the collision rate falling to $29.3\%$, and AvgD increasing to $5.13$~m. A single end-of-generation signal leaves the planner blind for most of the inference cycle, whereas the progressively aggregated latent keeps it steered by up-to-date intent, driving the consistent gains in both task completion and social interaction.

\begin{table}[t]
\centering
\caption{\textbf{Knowledge transfer strategy ablation} on the Realistic Dynamic Environment.
Both methods share the same slow VLM reasoner and fast expert planner; only the knowledge transfer mechanism differs.
\textbf{W-t-A} (Wait-then-Act): the fast planner blocks until the slow VLM completes full autoregressive generation, then receives a single conditioning signal from the final output;
\textbf{Stream} (Ours): the Token-Wise Hidden Streamer extracts intermediate hidden states as tokens are generated, and the Sequence-to-Slot Latent Bridge continuously updates the fast planner's conditioning, enabling it to act on progressively refined guidance while the slow VLM is still reasoning.
$\uparrow$ higher is better; $\downarrow$ lower is better.}
\label{tab:ablation_streaming}
\renewcommand{\arraystretch}{1.15}
\resizebox{\columnwidth}{!}{%
\begin{tabular}{l|ccc|ccc}
\toprule
& \multicolumn{3}{c|}{\textbf{Task Completion}}
& \multicolumn{3}{c}{\textbf{Social Interaction}} \\
\cmidrule(lr){2-4} \cmidrule(lr){5-7}
\textbf{Method}
& SR$\uparrow$ & NE$\downarrow$ & SPL$\uparrow$
& PSC$\uparrow$ & Col$\downarrow$ & AvgD$\uparrow$ \\
\midrule
W-t-A
& 24.80 & 7.45 & 21.64
& 90.9 & 29.7 & 4.97 \\
\rowcolor{sparkvln_blue!10}
\textbf{Stream}
& \textbf{34.80} & \textbf{6.21} & \textbf{28.68}
& \textbf{93.9} & \textbf{29.3} & \textbf{5.13} \\
\midrule
$\Delta$
& \gc{$+$10.0} & \gc{$-$1.24} & \gc{$+$7.04}
& \gc{$+$3.0} & \gc{$-$0.4} & \gc{$+$0.16} \\
\bottomrule
\end{tabular}%
}
\vspace{-2mm}
\end{table}

\begin{table}[t]
\centering
\caption{\textbf{Sequence-to-Slot Latent Bridge ablation} on the Realistic Dynamic Environment.
We compare two strategies for compressing the variable-length hidden stream into fixed-size planner conditioning.
\textbf{Mean-Pool}: collapse all streamed tokens into a single vector via global average pooling;
\textbf{Cross-Attn} (Ours): learnable queries attend to the projected stream via multi-head cross-attention, adaptively routing informative features into fixed-size latent slots.
$\uparrow$ higher is better; $\downarrow$ lower is better.}
\label{tab:ablation_meanpool}
\renewcommand{\arraystretch}{1.15}
\resizebox{\columnwidth}{!}{%
\begin{tabular}{l|ccc|ccc}
\toprule
& \multicolumn{3}{c|}{\textbf{Task Completion}}
& \multicolumn{3}{c}{\textbf{Social Interaction}} \\
\cmidrule(lr){2-4} \cmidrule(lr){5-7}
\textbf{Aggregation}
& SR$\uparrow$ & NE$\downarrow$ & SPL$\uparrow$
& PSC$\uparrow$ & Col$\downarrow$ & AvgD$\uparrow$ \\
\midrule
Mean-Pool
& 25.20 & 7.36 & 21.17
& 92.7 & 36.5 & 5.05 \\
\rowcolor{sparkvln_blue!10}
\textbf{Cross-Attn}
& \textbf{34.80} & \textbf{6.21} & \textbf{28.68}
& \textbf{93.9} & \textbf{29.3} & \textbf{5.13} \\
\midrule
$\Delta$
& \gc{$+$9.60} & \gc{$-$1.15} & \gc{$+$7.51}
& \gc{$+$1.2} & \gc{$-$7.2} & \gc{$+$0.08} \\
\bottomrule
\end{tabular}%
}
\vspace{-3mm}
\end{table}

\noindent\textbf{Bridge design.}
The Sequence-to-Slot Latent Bridge must compress a variable-length, high-dimensional hidden stream into a fixed-size conditioning matrix for the expert planner. We compare two aggregation strategies. \emph{Mean-Pool} collapses all streamed tokens into a single vector via global average pooling, discarding token-level granularity. \emph{Cross-Attn} (Ours) lets $N{=}8$ learnable queries attend to the projected stream via multi-head cross-attention, adaptively routing the most informative features into fixed-size latent slots. As shown in Tab.~\ref{tab:ablation_meanpool}, Cross-Attn improves SR from $25.20\%$ to $34.80\%$ ($+9.60$ points) and SPL from $21.17\%$ to $28.68\%$ ($+7.51$ points) over Mean-Pool, lowers NE from $7.36$~m to $6.21$~m, and improves every social metric (PSC $92.7\%$ to $93.9\%$, Col $36.5\%$ to $29.3\%$, AvgD $5.05$~m to $5.13$~m). This validates that naive pooling discards token-level structure, whereas our cross-attention bridge preserves and selectively aggregates it into a compact yet
expressive conditioning signal for the fast expert planner.

\begin{figure*}[t]
\centering
\includegraphics[width=0.82\linewidth]{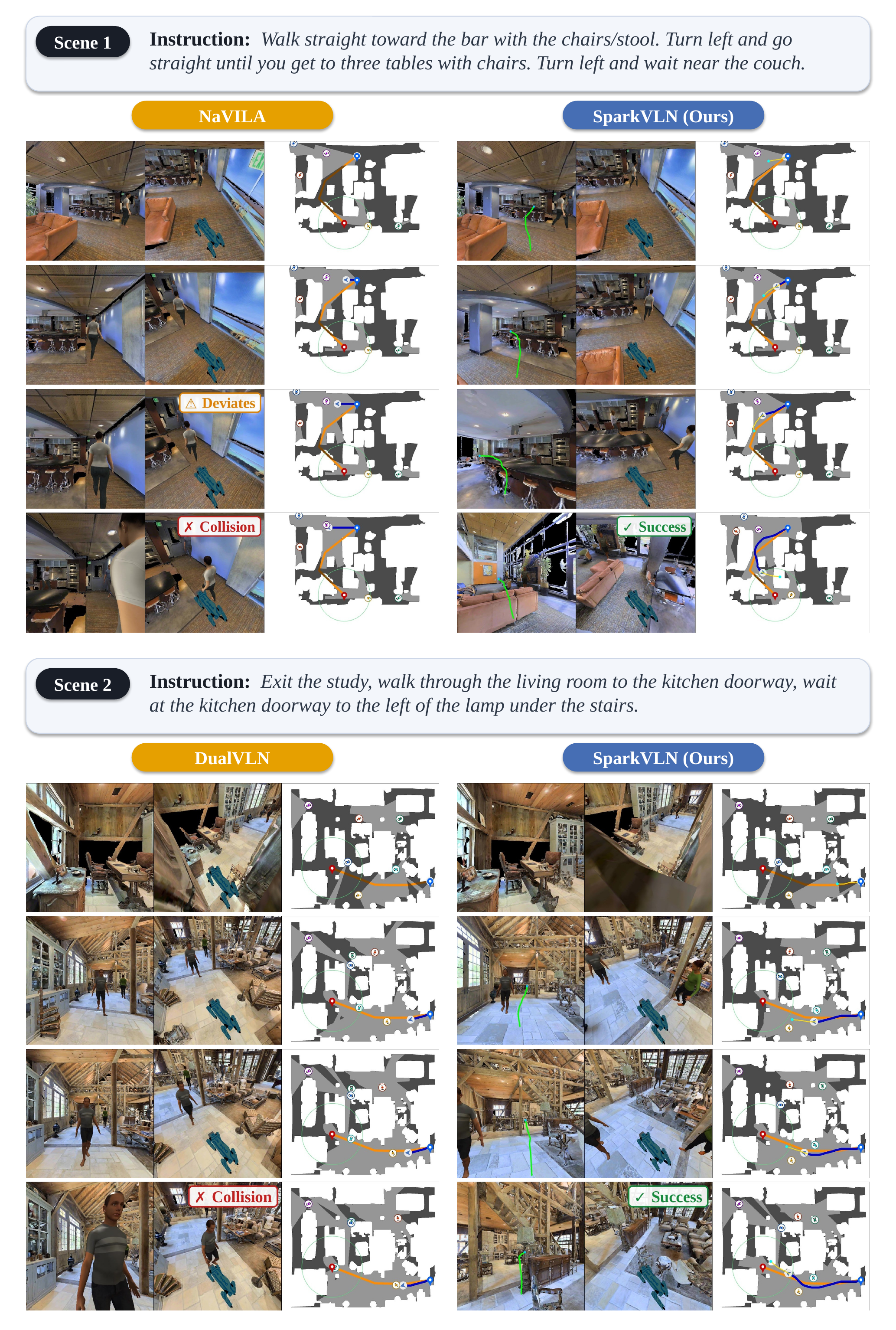}
\caption{\textbf{Qualitative comparison in dynamic environments.}
Each row shows the egocentric RGB view, the third-person view, and the top-down occupancy map, where the blue curve is the executed trajectory toward the goal (red pin) and the orange curve is the reference ground-truth path~(computed without pedestrians). All episodes are run in the \textit{Realistic Dynamic Environment}, where the simulator keeps running during inference and pedestrians advance while the model reasons. \textbf{Scene~1 (vs.\ NaVILA):} \sparkvln{} faithfully follows the instruction and proactively avoids the moving pedestrian, reaching the target near the couch, whereas NaVILA both deviates from the instructed route and eventually collides with a pedestrian. \textbf{Scene~2 (vs.\ DualVLN):} the dual-system baseline DualVLN must finish full VLM inference before transferring guidance to its fast planner, so the robot reacts late, colliding with the pedestrian in gray. \sparkvlnplain{} instead streams guidance during VLM decoding, smoothly negotiating both pedestrians and adhering to the instruction.}
\label{fig:vln_comparison}
\vspace{-4em}
\end{figure*}

\begin{figure*}[t]
\centering
\includegraphics[width=0.82\linewidth]{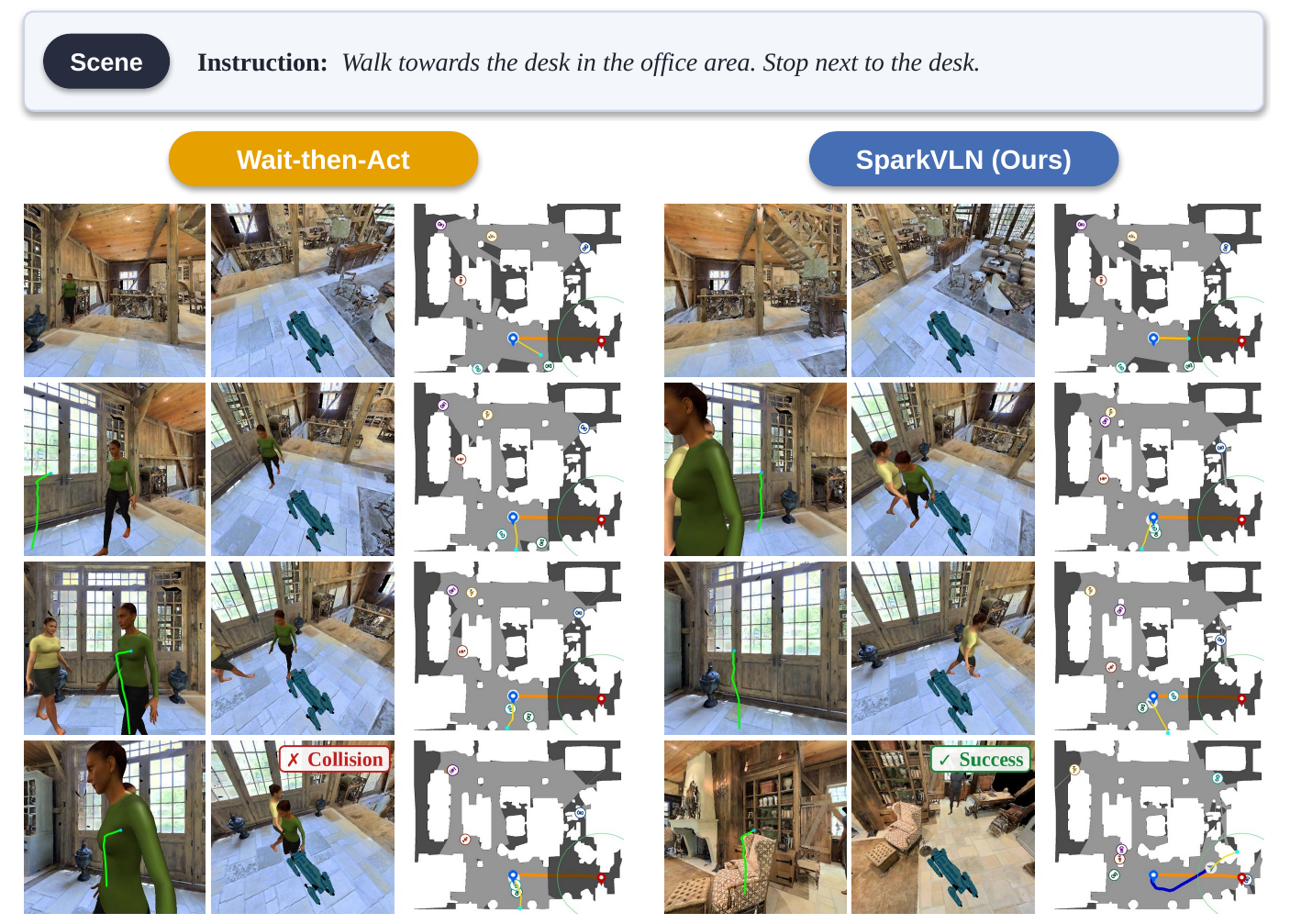}
\caption{\textbf{Qualitative comparison on a human-populated navigation episode.}
Given the same instruction (\textit{``Walk towards the desk in the office area. Stop next to the desk.''}), we compare the \textcolor[HTML]{E6A000}{\textbf{Wait-then-Act}} baseline (Left) with our \textcolor[HTML]{466EB4}{\textbf{\sparkvln{}}} (Right). Each row shows the egocentric RGB view, the third-person view, and the top-down occupancy map, where the blue curve is the executed trajectory toward the goal (red pin) and the orange curve is the reference ground-truth path~(computed without pedestrians). All episodes are run in the \textit{Realistic Dynamic Environment}, where the simulator keeps running during inference and pedestrians advance while the model reasons. The baseline stalls and then reacts abruptly, colliding with a moving pedestrian, whereas our method anticipates the human and follows the instruction to reach the target.}
\label{fig:chunk_ablation}
\end{figure*}

\noindent\textbf{Qualitative results.}
Fig.~\ref{fig:vln_comparison} and Fig.~\ref{fig:chunk_ablation} visualize representative episodes in the Realistic Dynamic Environment, where the simulator keeps running during inference and pedestrians advance while the model reasons. In Fig.~\ref{fig:vln_comparison}, \sparkvlnplain{} faithfully follows the instruction and proactively yields to the moving pedestrian, reaching the target specified by the language, whereas the single-system NaVILA drifts off the instructed route and eventually collides with a pedestrian. Against the dual-system baseline DualVLN, which must complete full VLM inference before issuing guidance to its planner, our model benefits from token-wise streaming and updates its plan far more frequently, negotiating both pedestrians smoothly instead of stalling and colliding. Fig.~\ref{fig:chunk_ablation} further compares the two transfer strategies. The Wait-then-Act baseline stalls during inference and then reacts abruptly, colliding with an oncoming pedestrian, while \sparkvlnplain{} anticipates the pedestrian from progressively refined guidance and reaches the goal. These cases mirror the quantitative gains in Tab.~\ref{tab:vln_3stage}, showing that streaming translates directly into safer, more compliant behavior under realistic latency.

\noindent\textbf{Runtime and staleness profile.}
Tab.~\ref{tab:runtime_profile} profiles per-update latency and the pedestrian displacement accrued during inference on a single RTX~4090 (batch size $=1$). Because streaming exposes usable guidance from the first token rather than only at end-of-generation, \sparkvlnplain{} (Stream) cuts the effective per-update latency from $0.788$~s (W-t-A) to $0.185$~s on average, the fastest among all VLM-based methods. Lower latency translates directly into less observation staleness, as the average pedestrian displacement during one inference drops from $0.213$~m (W-t-A) to $0.050$~m (Stream) and the worst case from $0.893$~m to $0.694$~m. In other words, low latency keeps the planner acting on up-to-date guidance, underpinning the safety and success gains reported above.

\begin{table}[t]
\centering
\setlength{\tabcolsep}{3.6pt}
\renewcommand{\arraystretch}{1.18}
\caption{\textbf{Runtime profile} in the Realistic Dynamic Environment.
We report per-inference latency and pedestrian motion during inference on a single RTX~4090 (batch size $=1$).
\textbf{Lat.}\ is the wall-clock time of one model forward (observation$\rightarrow$action);
\textbf{Disp.}\ is the pedestrian displacement (m) during that interval.
For both we report per-inference average (\textit{Avg}) and worst-case maximum (\textit{Max}).
The bottom block contrasts \textit{Wait-then-Act} (full-output blocking) with \textit{Streaming} (token-wise streaming).}
\label{tab:runtime_profile}
\resizebox{\columnwidth}{!}{%
\begin{tabular}{l|l|c|cc|cc}
\toprule
\multirow{2}{*}{\textbf{Method}}
& \multirow{2}{*}{\textbf{Backbone}}
& \multirow{2}{*}{\textbf{Params}}
& \multicolumn{2}{c|}{\textbf{Lat.\ (s)}}
& \multicolumn{2}{c}{\textbf{Disp.\ (m)}} \\
\cmidrule(lr){4-5} \cmidrule(lr){6-7}
& & & Avg$\downarrow$ & Max$\downarrow$ & Avg$\downarrow$ & Max$\downarrow$ \\
\midrule
NaVid~\cite{zhang2024navid}
& \raisebox{-0.3ex}{\includegraphics[height=1em]{figs/icons/llama-vid.png}}~LLaMA-VID~\cite{li2024llama}
& 7B   & 0.741 & 2.864 & 0.394 & 1.864 \\
Uni-NaVid~\cite{zhang2024uni}
& \raisebox{-0.3ex}{\includegraphics[height=1em]{figs/icons/llama-vid.png}}~LLaMA-VID~\cite{li2024llama}
& 7B   & 0.193 & \textbf{0.795} & 0.202 & 0.773 \\
NaVILA~\cite{cheng2025navila}
& \raisebox{-0.3ex}{\includegraphics[height=1em]{figs/icons/vila.png}}~VILA~\cite{lin2024vila}
& 8B   & 0.603 & 1.140 & 0.483 & 1.514 \\
DualVLN~\cite{wei2026ground}
& \raisebox{-0.3ex}{\includegraphics[height=1em]{figs/icons/qwen.png}}~Qwen-VL-2.5~\cite{Qwen2.5-VL}
& 7B   & 0.637 & 2.427 & 0.269 & 2.801 \\
\sparkvln~(W-t-A)
& \raisebox{-0.3ex}{\includegraphics[height=1em]{figs/icons/vila.png}}~VILA~\cite{lin2024vila}
& 8B & 0.788 & 1.325 & 0.213 & 0.893 \\
\rowcolor{sparkvln_blue!10}
\sparkvln~(Stream)
& \raisebox{-0.3ex}{\includegraphics[height=1em]{figs/icons/vila.png}}~VILA~\cite{lin2024vila}
& 8B & \textbf{0.185} & 1.030 & \textbf{0.050} & \textbf{0.694} \\
\bottomrule
\end{tabular}%
}
\vspace{-2em}
\end{table}

\section{Conclusion}
\label{sec:conclusion}
We presented \sparkvln{}, a dual-system framework for dynamic social VLN in which a slow VLM reasoner streams progressively aggregated reasoning latents to a fast flow-matching expert planner. Three coupling modules turn reasoning from a blocking call into a continuous signal: a Token-Wise Hidden Streamer extracts intermediate hidden states, a Sequence-to-Slot Latent Bridge projects them into fixed-size latent slots, and an Evolving Latent Conditioner infuses those slots into the fast planner, making fresh guidance available during reasoning rather than only at the end of inference. We also introduced an asynchronous, human-centric benchmark suite that keeps pedestrians and the robot active during inference and reports task-completion, social-interaction, and inference speed statistics. Across both the Idealized and Realistic Dynamic Environments, \sparkvlnplain{} improves navigation success and social compliance while cutting inference latency. Future work includes extending to longer-horizon instructions with multiple sub-goals and incorporating pedestrian intent prediction for anticipatory, rather than reactive, social planning.


\bibliographystyle{IEEEtran}
\bibliography{IEEEabrv,ref}

\end{document}